\pdfoutput=1

\documentclass[11pt]{article}

\usepackage[final]{coling}

\usepackage{times}
\usepackage{latexsym}

\usepackage[T1]{fontenc}

\usepackage[utf8]{inputenc}

\usepackage{microtype}

\usepackage{inconsolata}

\usepackage{graphicx}

\usepackage{pifont}
\newcommand{\cmark}{\ding{51}}%
\newcommand{\xmark}{\ding{55}}%
\usepackage{amsmath}
\usepackage{graphicx}
\usepackage{booktabs}
\usepackage{tikz}
\usepackage{pgfplots} 
\usepackage{scalefnt} 
\usepackage{caption}
\usepackage{subcaption}
\usepackage{enumitem}
\usepackage{todonotes}
\usepackage{colortbl}
\usepackage{makecell}
\usepackage{multirow}
\usepackage{xcolor}         
\usepackage{spverbatim}

\newcommand{\mytask}{Case2Code}

\title{Case2Code: Scalable Synthetic Data for Code Generation}

\author{
    Yunfan Shao$^{1,2}$,
    Linyang Li$^{2\dag}$,
    Yichuan Ma$^{1,2}$,
    Peiji Li$^{1,2}$,
    Demin Song$^{2}$, \\ \bf
    Qinyuan Cheng$^{1,2}$,
    Shimin Li$^{1}$,
    Xiaonan Li$^{1}$,
    Pengyu Wang$^{1}$,
    Qipeng Guo$^{2}$, \\ \bf
    Hang Yan$^{2,3}$, 
    \bf Xipeng Qiu$^{1\dag}$,
    Xuanjing Huang$^{1}$,
    Dahua Lin$^{2,3}$ \\
    $^1$School of Computer Science, Fudan University\\
    $^2$Shanghai AI Laboratory\\
    $^3$The Chinese University of Hong Kong \\
    \texttt{\{yfshao19, xpqiu\}@fudan.edu.cn} \\
    \texttt{\{lilinyang, yanhang\}@pjlab.org.cn}
}

\begin{document}
\maketitle
\def\thefootnote{$\dag$}\footnotetext{Corresponding Authors.}\def\thefootnote{\arabic{footnote}}

\begin{abstract}

Large Language Models (LLMs) have shown outstanding breakthroughs in code generation. 
Recent work improves code LLMs by training on synthetic data generated by some powerful LLMs, which can be challenging to scale due to the dependence on a teacher model and high generation costs.
In this paper, we focus on synthesizing code data at scale and propose a \textbf{Case2Code} task by exploiting the expressiveness and correctness of programs. \textbf{Case2Code} is an inductive inference task that aims to infer underlying code implementations by observing input-output examples or program behaviors, By incorporating LLMs to generate program inputs, and executing the program with these inputs to obtain the program outputs, we can synthesize diverse and high-quality \textbf{Case2Code} data at scale for training and evaluating code LLMs.
Experimental results show that case-to-code induction is challenging for current representative LLMs if they are untrained. Models trained with \textbf{Case2Code} improve performance not only on distribution case-to-code induction but also on various coding-generation tasks, demonstrating the great potential of large-scale synthetic data and inductive learning.\footnote{Code and datasets are available at \url{https://github.com/choosewhatulike/case2code}.}

\end{abstract}

\section{Introduction}
The success of large language models (LLMs), exemplified by GPT-4~\cite{2023gpt4} has revolutionized the AI community.
One of the most impressive abilities of LLMs is the code generation ability, exemplified by writing code blocks or programs to accomplish complex instructions and satisfy user specifications~\cite{hong2024metagpt,zhuo2024bigcodebench,2024codesurvey}. 

To further improve the performance of code generation with open-source LLMs, recent work adopts synthetic data via self-instruct~\cite{2023selfinstruct,alpaca} to distill the strong capability of code generation from teacher models. Specifically, practitioners often collect and devise code instructions and then generate high-quality responses using a powerful teacher LLM to synthesize the training data. Then, the data can be used to fine-tune a weak LLM to bootstrap their code generation ability. 
For example, Code Alpaca~\cite{codealpaca} incorporates ChatGPT to generate 20K code instructions under 21 seed tasks for fine-tuning.
WizardCoder~\cite{luo2023wizardcoder} adopts Evol-Instruct to the realm of code, using powerful closed-source LLMs to synthetic high-quality code instruction-following samples to enhance open-sourced LLMs code generation performance.

\begin{figure}
    \centering
    \includegraphics[width=0.5\textwidth]{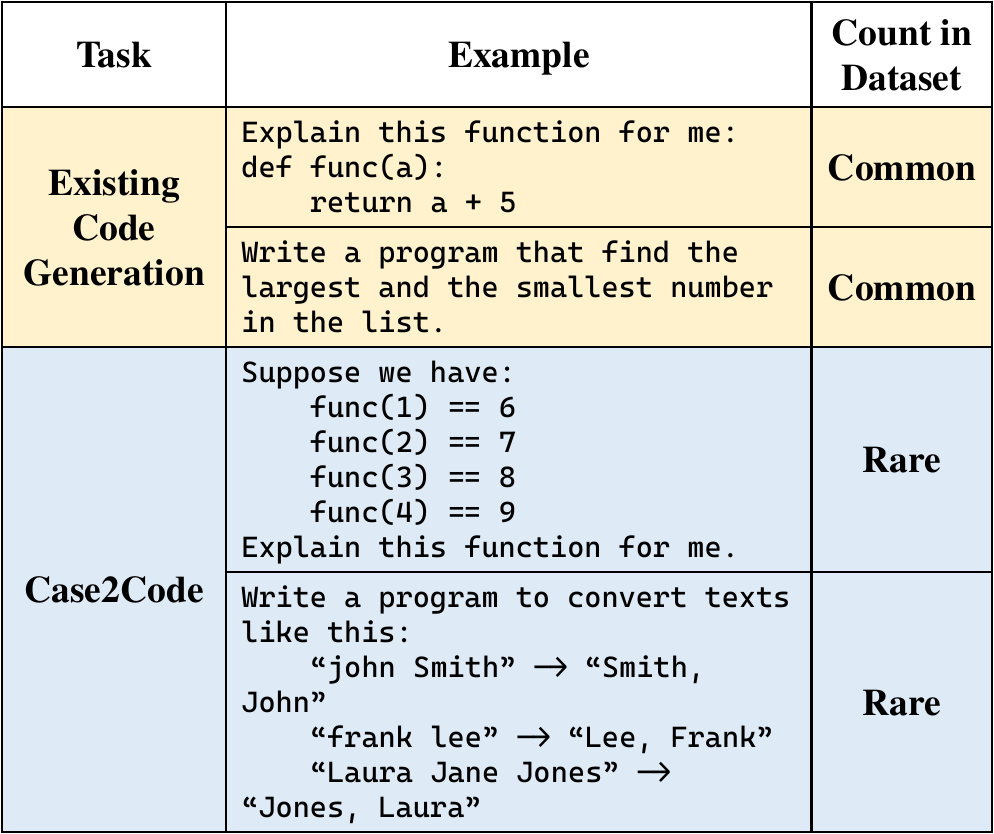}
    \caption{Examples of existing code generation and case2code tasks. Compared with existing code generation instructions, inductive learning tasks like case2code are rare in the training data, which makes it challenging for LLMs to perform.
    }
    \label{fig:reasoning-example}
\end{figure}

Despite the success of these code synthetic data, they rely on a powerful teacher for data generation~\cite{yu2023bias}, which can be challenging to scale up as they are bound by the capability of the teacher and suffer from high costs. Moreover, models can perform poorly on tasks like inductive learning as these instructions are rare in the training corpus (Figure~\ref{fig:reasoning-example}). 
Therefore, in this paper, to synthesize high-quality code data at scale and complement existing code training data, we introduce \textbf{Case2Code}, a diverse and challenging synthetic task for LLMs.

Inspired by inductive inference tasks~\cite{balog2016deepcoder,devlin2017robustfill,ellis2021dreamcoder}, we focus on large-scale data synthesis with programs in the real world. In Case2Code, samples are synthesized from real-world productive functions, which are closer to the actual distribution of general LLM applications and production.
Specifically, the Case2Code challenge requires the LLM to infer the underlying program based on several input-to-output cases generated by the real-world program.
In Case2Code learning, LLMs are supposed to write solutions formulated by codes based on the example outputs, 
which is one common scenario in the real-world working process, using examples to convey knowledge.

To obtain large-scale and diverse Case2Code data, we first gather a diverse collection of executable code texts that cover a wide range of real-world applications.
Then, we generate the input-output transformation cases with the assistance of LLMs and code interpreters. 
By incorporating LLMs to write input examples for each program and execute the program with these inputs to gather the corresponding outputs, we can synthesize large-scale \mytask~samples with diverse data transformations and complicated control logic. The data synthetic framework does not require powerful LLMs with advanced code generation capabilities, resulting in the possibility of a weak-to-strong learning paradigm.

Based on the synthetic data, we can form a unique and challenging task to evaluate and further train the LLMs and study the case-to-code induction ability of LLMs.
In the Case2Code challenge, we first test how current LLMs perform in making the code induction task. 
We then train LLMs with Case2Code data to further study whether such data can improve the in-distribution code induction ability and generalize to other commonly used code generation tasks.
Experimental results show that Case2Code is a challenging task for LLMs, even for powerful LLMs like LLaMA3-70B, GPT-3.5, and GPT-4.
With constructed Case2Code data, we can boost LLMs to learn to make such inductive inference tasks, while such ability can be transferred to help improve general code generation tasks such as HumanEval and MBPP.

To summarize, in this paper, we:

    (1) We introduce a scalable data synthetic framework that aims to generate high-quality and diverse inductive code generation samples for evaluating and training LLMs, \textbf{Case2Code}.
   
    (2) We evaluate the inductive learning ability of current representative LLMs, demonstrating the necessity of synthesizing inductive data like \textbf{Case2Code} for LLMs.
  
    (3) We explore methods of training LLMs on large-scale \textbf{Case2Code} data, showing not only great improvements on the Case2Code challenge but also a consistent generalization of the trained models in general code generation tasks.


\section{Related Work}

Our work discusses the reasoning ability of LLMs, touching on the following grounds:

\subsection{Inductive Inference}
Inductive inference is rarely discussed in LLM reasoning, most research focuses on specific scenarios with limited inductive reasoning.
One pioneer work is prerequisite toy tasks \cite{weston2015towards} where the task goal is to solve simple induction.
Later, \citet{yang2022language} introduces various world-wide knowledge such as botany, history, and geography into the facts given and asks neural models to predict whether a given rule is correct.
In the realm of code, several works focus on Programming by Example (PBE), which aims to induce a valid program given the expected inputs and the corresponding outputs. These works train and evaluate inductive program synthesis models for constrained scenarios with limited search spaces, such as operations on list, string, and manually-defined objects~\cite{balog2016deepcoder,devlin2017robustfill,ellis2021dreamcoder,shi2023exedec,WuWJCRX24,keven2024pbellm}.
Different from previous studies, our proposed Case2Code task leverages diverse code in the real world as a powerful platform for LLMs to learn inductive inference under various challenging scenarios.

\begin{figure*}[!htbp]
    \centering
    \includegraphics[width=0.95\textwidth]{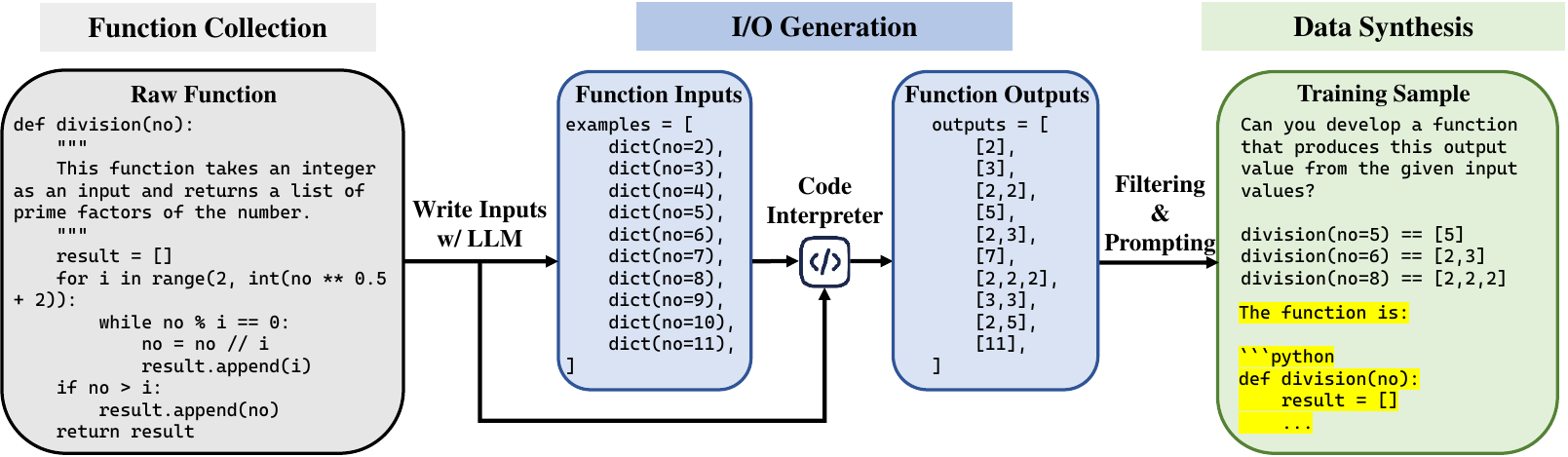}
    \caption{Our synthetic framework incorporates an LLM and a code interpreter to construct \mytask~training samples at scale automatically.}
    \label{fig:overview}
    
\end{figure*}

\subsection{Synthetic Data}

Recent works focus on building high-quality instruction-following or question-answering training data through strong LLMs such as GPT-4 to enhance smaller LLMs~\cite{yu2023metamath,ms2024orcamath,luo2023wizardcoder}.
While a particular line of work focuses on studying different strategies to diverse the instructions and control the quality of LLM generation, including self-consistency~\cite{wang2022self}, rejection sampling~\cite{huang2023selfimprove,2023rftmath,wang2023mathshepherd}, program-of-thought (PoT)~\cite{2024codesurvey}, tree-structure CoT (ToT) searching \cite{Yao2023TreeOT}, Monte Carlo Tree Searching \cite{google2016alphago,Chen2024AlphaMathAZ}, etc. These methods still require a strong LLM as the teacher with high costs of model inference, limiting the scalability.

\section{Method}
In this section, we illustrate the framework for synthesizing \mytask~data in detail, which focuses on producing large-scale and high-quality inductive reasoning data in the code domain.
Unlike other synthetic data frameworks that distill high-quality training data from a strong teacher LLM to provide supervision signals to improve student LLMs, our \mytask~synthetic framework introduces a writer LLM to assist the synthesis of data samples. Thus the overall data quality does not directly rely on the performance of the LLM generator. And we can efficiently obtain reliable \mytask~training data at scale.

\subsection{Problem Formulation}
The inductive reasoning task aims to find a general hypothesis based on a small set of observations to explain a phenomenon. In this paper, we define \mytask, an inductive reasoning task in the code domain.
\mytask~is a program synthesis task that targets the reconstruction of unknown programs based on observations of the program behaviors.

Formally, for a functional program $\mathcal{P}$, we have a set of $n$ input-output examples $\mathcal{S_P}=\{(x_1,y_1), (x_2,y_2), ..., (x_n, y_n)\}$, where $y_i = \mathcal{P}(x_i), i = 1, 2, ..., n$. The goal of \mytask~is to implement a program $\mathcal{P'}$ that captures the functionality of the program $\mathcal{P}$ based on the observed set of input-output example cases $\mathcal{S_P}$. And for any new input case $x_{\text{new}} \notin \mathcal{S_P}$, the implemented program $\mathcal{P'}$ should satisfy that $\mathcal{P}(x_{\text{new}})=\mathcal{P'}(x_{\text{new}})$.

\subsection{Framework Overview}
In our synthetic data generation framework, we focus on generating large-scale and diverse \mytask~data automatically. As shown in Figure~\ref{fig:overview}, we first collect diverse programs from large-scale datasets with rule-based filters. Then we incorporate LLMs to write diverse example inputs and utilize the code interpreter to calculate their corresponding outputs for each program. Finally, we filter out low-quality programs based on their outputs and convert the obtained triple (program, inputs, outputs) into \mytask~data for inductive reasoning in the code domain. 

Note that the correctness of our synthetic data does not depend on the capabilities of the used LLMs. Therefore, we can synthetic high-quality \mytask~data at scale using small LLMs with low costs. 

\subsection{Collecting Programs}
To obtain massive data samples for inductive reasoning learning, we first need to acquire massive and diverse programs that take input arguments, do some complicated processes, and return output values. 
Instead of prompting LLMs to generate functions that meet these requirements, we collect human-written high-quality programs in the wild to enhance diversity. 

Specifically, we sample valid Python functions from The Stack~\cite{Kocetkov2022TheStack} to construct our reasoning dataset. We incorporate the out-of-box Abstract Syntax Tree (AST) parsing tool~\footnote{https://docs.python.org/3/library/ast.html} to parse each file in The Stack to obtain Python functions. We only keep self-contained high-quality functions that satisfy all of these filtering rules: (1) pass the syntax check; (2) have one or more input arguments and return values; and (3) do not rely on third-party packages or external I/O operations. After collecting these functions, we can easily execute and verify these functions to obtain diverse \mytask~data with a simple and fast code interpreter at scale, which avoids extra file or network operations that require a sophisticated sandbox.

\subsection{Generating Inputs}

Once we collect large-scale functions, the next step is to obtain the corresponding input-output pairs for each function to construct the \mytask~data. It is infeasible to write test cases for each function manually. So, we utilize LLMs to generate suitable input examples for these functions. We prompt LLMs to write some example input arguments for each function based on the corresponding function implementation. Detailed prompt is listed in Table~\ref{tab:appendix-generator-prompt} in the appendix.

To generate suitable input arguments, the LLM needs first to analyze the implementation of the functions, then infer the possible types and value ranges of the input arguments, and finally come up with correct input arguments. However, we argue that a powerful LLM is not the key factor for our synthetic data. As we find that while strong LLMs can write high-quality inputs to generate \mytask~training data that boosts the reasoning performance of weak LLMs, the weak LLM can also write inputs for creating \mytask~data to self-improve their reasoning ability (see Sec~\ref{sec:generator-ablation}). Therefore, the generation process can be scaled efficiently at a low cost by using small LLMs.

\subsection{Obtain Outputs}
After collecting self-contained functions and the corresponding inputs, it is intuitive to incorporate a code interpreter to run these functions on their inputs for output curation. Since the LLM-generated input examples can contain errors, we introduced a filtering procedure to reject invalid inputs or functions based on their returned outputs. Specifically, if the outputs of a function do not change as the inputs change (e.g. always return the same output or exceptions), the function is considered invalid and will be filtered out. 

Moreover, we also filter out functions that generate very long output values to ensure the length of the generated \mytask~data is within the context window size of current LLMs. Note that we do not filter out inputs that lead to exceptions or runtime errors, as we believe that failure call attempts can also provide valuable information for inductive reasoning to reconstruct the function.

\subsection{Post-processing}
The final step is to convert the obtained functions and their corresponding input-output pairs into \mytask~style data. Formally, for a given function $\mathcal{P}$ and its $n$ test cases $\mathcal{S_P}=\{(x_1,y_1), (x_2,y_2), ..., (x_n, y_n)\}$, we randomly sample $m$ examples ($m <= n$) as the observed set $\mathcal{S'_P}$. We generate the prompted data that facilitate the LLM to conduct inductive reasoning on the observed examples $\mathcal{S'_P}$ to reconstruct the given function $\mathcal{P}$. Converted training examples are shown in Table~\ref{tab:appendix-data-examples} in the appendix. 

We find that the diversity of the prompts can substantially affect the generalization of the model reasoning performance (as shown in Sec~\ref{sec:prompt-ablation}). Therefore, we manually construct about 10 prompts with different styles to enhance the data diversity.

\section{Experiment}
In this section, we illustrate the experimental setups and discuss the experimental results to demonstrate the challenge of solving \mytask~problems and show the effectiveness of large-scale \mytask~synthetic data.
\begin{table*}[htbp]
  \centering
    \begin{tabular}{lc|cccc|c}
    \toprule
    & \textbf{Size} & \textbf{HumanEval} & \textbf{HumanEval}+ & \textbf{MBPP} & \textbf{MBPP}+ & \textbf{Case2Code} \\
    \midrule
    GPT-4 & - &90.2 & 86.6 & 85.7 & 73.3 & 43.6 \\
    GPT-3.5 & - & 76.8 & 70.7 & 82.5 & 69.7 & 34.2 \\

    \midrule
    \multirow{4}{*}{LLaMA2-Chat} & 7B & 14.0  & 11.6  & 26.8 &  20.3 & 0.2 \\
        & 13B & 23.1 & 19.5  & 37.0  &  27.6 & 8.2 \\
        & 34B & 22.6 & - & 33.0 & - & - \\
        & 70B & 36.6 & 28.7  & 46.3  & 35.1  & 7.8 \\
    \midrule
    \multirow{3}{*}{CodeLLaMA-Instruct}
        & 7B & 37.8 & 35.4 & 59.5 & 46.8 & 14.2 \\
        & 13B & 42.7 & 38.4 & 63.5 & 52.6 & 19.0 \\
        & 34B & 51.8 & 43.9 & 69.3 & 56.3 & 22.6 \\
    \midrule
    \multirow{2}{*}{LLaMA3-Instruct} & 8B & 61.6 & 56.7 & 70.1 & 59.3 & 23.2 \\
        & 70B &77.4 & 72.0 & 82.3 & 69 & 34.0 \\
    \bottomrule
        \end{tabular}%
  \caption{Accuracy of various representative LLMs on the code generation datasets and the \mytask~test set.}
  \label{tab:zeroshot-result}%
\end{table*}%

\subsection{Experimental Setup}
\paragraph{Data Construction}
We randomly sampled about 2.3 million functions from The Stack pre-training dataset, in which we already performed data deduplication with the evaluation benchmarks (e.g. HumanEval, MBPP, etc). We conduct the data synthetic pipeline incorporating InternLM2-7b~\cite{2024internlm2} to generate input examples for each function. The temperature is set to 0.2 and the top\_p is set to 0.95. The generation takes about 500 GPU hours using A800 GPUs. Then we use 64 CPUs to execute and filter functions, which takes about 1 hour. The execution is under a constrained Python environment to ensure safety. We eventually obtained 1.3M high-quality functions with input-output pairs for \mytask~reasoning. We hold out 500 samples for evaluation and the rest for training. 
For the hold-out evaluation samples, we further prompted GPT-4 (\texttt{gpt-4-turbo-2024-04-09}) to generate additional input examples and collect the corresponding outputs for a more strict inductive reasoning evaluation.

\paragraph{Training Setup}
To demonstrate the generalization and effectiveness of our synthetic training data, we conduct three variants of \mytask~training: direct fine-tuning, mixed pre-training, and mixed fine-tuning.
All \mytask~variants are trained for 5k steps with a batch size of 64, a maximum context window size of 4096, and apply linear warmup and cosine decay of the learning rate from the peak value of 2e-5 to 5e-6. All model training is completed on two servers of eight A800 GPUs. We conduct training on open-sourced models, i.e. InternLM2-7B~\cite{2024internlm2} and LLaMA3-8B~\cite{llama3modelcard} to verify the effectiveness of synthetic training data on different model series. 

\paragraph{Evaluation Setup}
We evaluate the coding ability of trained LLMs with HumanEval, MBPP. To conduct strict evaluation, we use EvalPlus, an extension to the original HumanEval and MBPP with massive additional test cases. For models that are not instructed tuned, we apply zero-shot prompting and four-shot prompting for HumanEval and MBPP evaluation, respectively. And for instructed-aligned LLMs, we use zero-shot prompting on all these benchmarks. To evaluate inductive reasoning on code, we test various LLMs on solving \mytask~tasks, with zero-shot prompting. When evaluating the instructed models that are not tuned on \mytask~task, we find the performance is unstable and sensitive to the prompts. We manually optimized the prompts for \mytask~evaluation to elicit the actual inductive reasoning ability of these models. We use greedy decoding during the inference for all experiments. 

\paragraph{Models}
We compare the trained models with several families of representative LLMs: GPT series\cite{2023gpt4}, CodeLLaMA~\cite{rozi2023codellama}, LLaMA2~\cite{hugo2023llama2} and LLaMA3~\cite{llama3modelcard}. For GPT series, we evaluate GPT-3.5 (\texttt{gpt-3.5-turbo-0125}) and GPT-4 (\texttt{gpt-4-turbo-2024-04-09}). For other model series, we evaluate their available open-sourced versions.

\begin{table*}[htbp]
  \centering
  \resizebox{0.95\textwidth}{!}{
    \begin{tabular}{lc|cccc|c}
    \toprule
    & \multirow{2}{*}{\parbox{1.5cm}{\textbf{Train w/ \\ Ours}}} & \multirow{2}{*}{\textbf{HumanEval}} & \multirow{2}{*}{\textbf{HumanEval}+} & \multirow{2}{*}{\textbf{MBPP}} & \multirow{2}{*}{\textbf{MBPP}+} & \multirow{2}{*}{\textbf{\mytask}} \\
    & & & & & & \\
    \midrule
    InternLM2-7B-Base & \xmark & 31.1 & 21.3 & 51.4 & 40.3 & 27.2$^\dag$ \\
    w/ Direct Fine-tuning & \cmark & \textbf{44.5} & 34.8 & 56.0 & 40.4 & \textbf{44.4} \\
    w/ Mixed Pre-training & \cmark & 43.9 & \textbf{40.9} & \textbf{58.4} & \textbf{42.6} & 41.4 \\
    \midrule
    InternLM2-7B & \xmark & 39.0 & 33.4 & 56.8 & \textbf{54.1} & 25.6$^\dag$ \\
    w/ Direct Fine-tuning & \cmark & 43.3 & 40.9 & 54.5 & 40.6 & \textbf{44.5}\\
    w/ Mixed Pre-training & \cmark & 47.6 & 37.2 & 58.4 & 45.6 & 42.4 \\
    w/ Insturction-tuning & \xmark & 49.4 & 43.9 & 58.0 & 50.4 & 6.2 \\
    w/ Mixed Instruction-tuning & \cmark & \textbf{64.6} & \textbf{56.7} & \textbf{63.4} & 52.4 & 44.0 \\
    \midrule
    LLaMA3-8B & \xmark & 35.4 & 20.1 & 59.1 & 45.1 & 29.2$^\dag$ \\
    w/ Direct Fine-tuning & \cmark & 43.2 & 39.0 & 50.6 & 35.1 & 44.8 \\
    w/ Mixed Pre-training & \cmark & 47.6 & 40.9 & 55.6 & 41.1 & 42.6 \\
    w/ Insturction-tuning & \xmark & 49.8 & 45.7 & 57.6 & 47.9 & 8.6 \\
    w/ Mixed Instruction-tuning & \cmark & \textbf{64.8} & \textbf{57.9} & \textbf{71.2} & \textbf{53.1} & \textbf{45.0} \\
    \bottomrule
    \end{tabular}%
    }
  \caption{Results of models trained with our synthetic dataset and the corresponding generalization performance. \mytask~performance are evaluated with zero-shot prompting, except results with $^\dag$, which are evaluated with four-shot prompting.}
  \label{tab:learn-result}%
  
\end{table*}%

\subsection{Zero-shot \mytask~is Challenging for Current LLMs}
As shown in Table~\ref{tab:zeroshot-result}, we report the zero-shot \mytask~performance of different representative LLMs and their programming performance. We can find that the zero-shot \mytask~performance of representative models is strongly related to their corresponding program synthesis performance. Models with higher program synthesis scores tend to achieve higher \mytask~performance. And larger models often outperform small models. This indicates that \mytask~can become a good benchmark to reflect the code reasoning performance of different LLMs. However, the zero-shot \mytask~scores of LLMs have a large gap compared with their coding accuracy, which demonstrates that existing LLMs are better at some types of reasoning (e.g. writing programs based on instructions) than others (e.g. inductive programs by their behaviors). This can be explained as the LLMs are trained with massive program generation data but fewer samples similar to \mytask~that need inductive reasoning. Similar to the Reverse Curse~\cite{berg2023ReverseCurse}, models trained with deductive reasoning data struggle to transfer to inductive reasoning tasks.

\subsection{Generalization of Training on \mytask}
One essential issue of synthetic data is its generalization ability. Therefore, we train different LLMs with our synthetic \mytask~dataset under various settings to explore how it affects the learning of code reasoning of LLMs. 

\subsubsection{Direct Fine-tuning}
First, we find that LLMs that are directly trained on the \mytask~reasoning samples can effectively learn coding based on cases. As shown in Table~\ref{tab:learn-result}, by direct fine-tuning, Internlm2-7B and LLaMA3-8B can significantly outperform the few-shot prompting baselines by up to 18.9\%, achieve up to 44.5\% and 42.0\% accuracy on \mytask~evaluation set, respectively, which even outperforms the more powerful LLMs like LLaMA3-70B, GPT-3.5, and comparable with GPT-4 (results in Table~\ref{tab:zeroshot-result}). Moreover, models trained with \mytask~reasoning also improve their program synthesis performance on benchmarks like HumanEval and MBPP. This indicates that the \mytask~reasoning is general and challenging. Training on \mytask~samples not only boosts the inductive reasoning performance in distribution but enhances the code understanding and code generation abilities of LLMs. As the \mytask~samples can be synthetic at scale, we believe that synthesizing large-scale and high-quality inductive reasoning data is a promising path to consistently improve LLMs without exhausting data. 

\subsubsection{Mixed Training}
\label{sec:mixed-training}
Then, we explore how to better incorporate our synthetic \mytask~data into different stages of LLM training to enhance the reasoning ability of LLMs in general. Specifically, we train LLMs with two variants of data mixing, either during pre-training or in the supervised fine-tuning (SFT) stage. The first mixing strategy introduces natural language pre-training texts from the Pile~\cite{gao2021pile} and the code pre-training samples from The Stack~\cite{Kocetkov2022TheStack}. The mixing ratio is 1:1:2 for samples from the Pile, The Stack, and the \mytask~dataset, respectively. On the other hand, we incorporate a supervised fine-tuning (SFT) dataset from WizardCoder~\cite{luo2023wizardcoder} to demonstrate that the performance gain of \mytask~training does not come from the understanding of instructions but the learning of inductive reasoning of code execution. We combine the SFT dataset with \mytask~samples in a 1:3 ratio, as the size of our synthetic dataset is much larger.

\paragraph{Mixed Pre-training}
As shown in Table~\ref{tab:learn-result}, when incorporated into the pre-training stage, the \mytask~training data helps the model to connect the execution states with the function implementation, which further facilitates the program synthesis performance of these LLMs. Compared with directly fine-tuned on \mytask~dataset, training these samples with pre-training texts enables the generalization of inductive reasoning of code states learned by the \mytask~task.

\paragraph{Mixed Instruction-tuning}
When trained with instruction-following datasets, the \mytask~data also improves the performance of the programming with instruction tasks, as reported in Table~\ref{tab:learn-result}. We evaluate the SFT models with the zero-shot instructed version of programming synthesis tasks, HumanEval, and MBPP. We find that incorporating \mytask~data boosts the performance of various LLMs on code generation tasks. Compared to the corresponding SFT baselines, InternLM2-7B improves on HumanEval from 49.4\% to 64.6\%, with more than 10\% improvements. LLaMA3-8B achieves 64.6\%, 57.9\%, and 71.2\% on HumanEval, HumanEval+, and MBPP, respectively, with significant improvements compared to the SFT version. These results demonstrate the effectiveness of learning on \mytask~and the necessity of incorporating inductive reasoning data into LLM training.

\subsection{Ablation Study}
In this section, we conduct ablation studies to demonstrate the effectiveness of the \mytask~synthetic pipeline across different families and scales of LLMs.

\begin{figure}[htbp]
    \centering
    \includegraphics[width=0.48\textwidth]{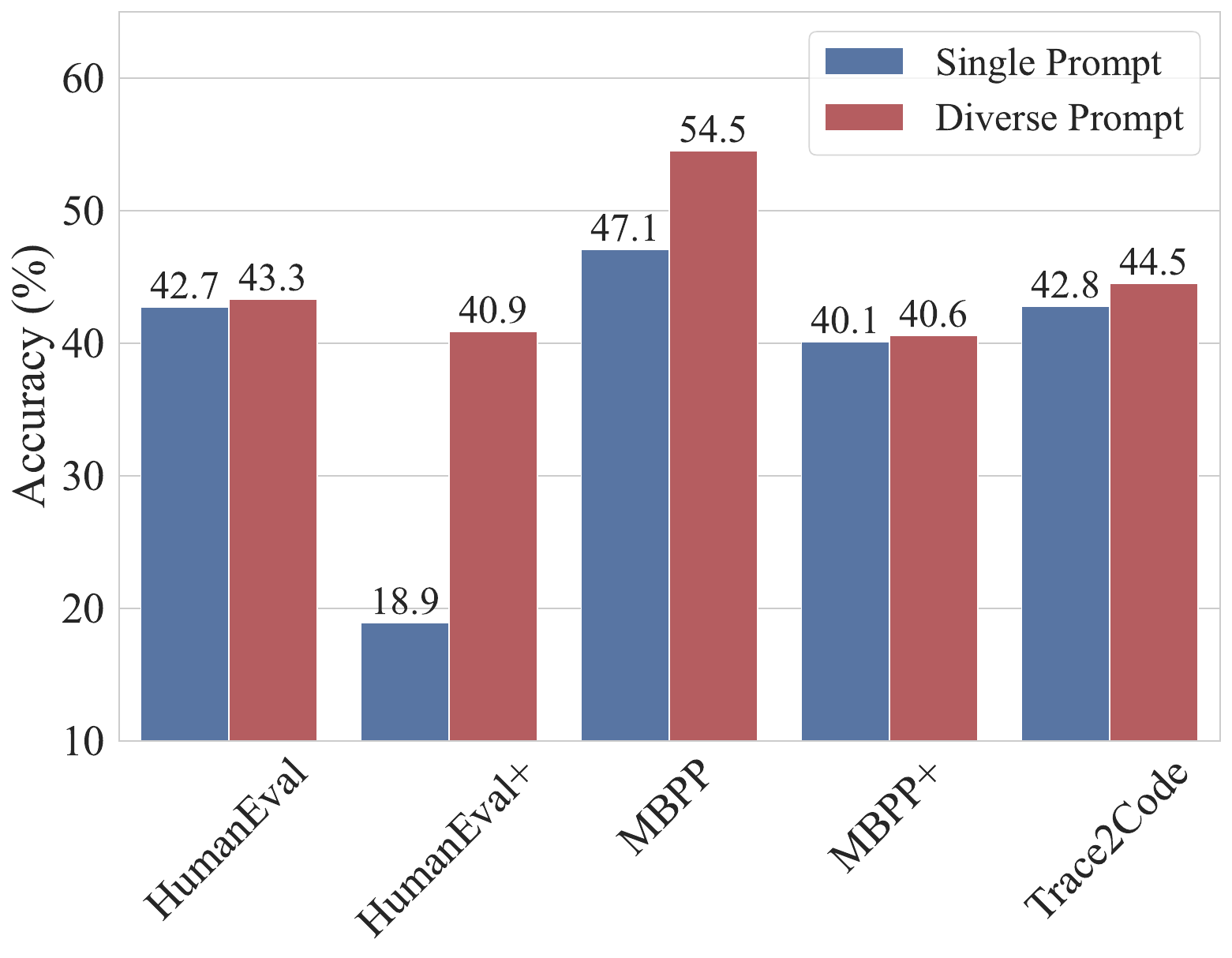}
    \caption{Downstream results when directly fine-tuning InternLM2-7B with different \mytask~prompt templates. Diverse prompts not only help the model to learn \mytask~reasoning but also significantly advance the generalization of the code inductive reasoning.}
    \label{fig:prompt-ablation}
\end{figure}

\paragraph{Prompt Diversity of Training Data}
\label{sec:prompt-ablation}
Since the synthetic \mytask~training data is converted by triples of (\textit{\textbf{programs}}, \textit{\textbf{inputs}}, \textit{\textbf{outputs}}), during the construction, the prompt templates are utilized to embed the input-output pairs to form natural language texts for LLM to learn. As the LLM can only rely on these converted prompts to learn the \mytask, it is important to understand the effectiveness of how different prompt templates affect the training of LLMs. Intuitively, the diversity of prompt templates plays an important role in the learning of LLMs. Therefore, we compare synthetic data prompted using a single template style with data utilizing diverse styles of templates. The result is reported in Figure~\ref{fig:prompt-ablation}, in which diverse prompts may have little effect on the in-domain \mytask~performance, however, the diversity significantly affects the accuracy of LLMs on out-of-domain program synthesis tasks. It is indicated that diversity can be critical during LLM learning, which also has been discussed in other domains like in general natural language processing tasks~\cite{jason2022flan} and alignment~\cite{openai2022instructgpt,2023selfinstruct}.
\begin{table}[htbp]
  \centering
  \resizebox{0.48\textwidth}{!}{
    \begin{tabular}{l|ccc}
    \toprule
      & HumanEval & HumanEval+ & Case2Code \\
    \midrule
    LLaMA3-8B & 35.4 & 20.1 & 29.2 \\
    \midrule
    w/ \mytask~(full) & \textbf{47.6} & \textbf{40.9} & \textbf{42.6} \\
    w/ \mytask~(code only) & 38.41 & 28.66 & 28.2 \\
    \bottomrule
    \end{tabular}%
    }
  \caption{Models trained with \mytask~(Full) achieve higher accuracy on multiple datasets than models trained on \mytask~(Code only), indicating the effectiveness of combining the I/O messages for \mytask~learning.}
  \label{tab:data-ablation}%
\end{table}%

\begin{table}[htbp]
  \centering
  \resizebox{0.48\textwidth}{!}{
    \begin{tabular}{l|cccc}
    \toprule
      & TP & TGS & Costs & \# Samples \\
    \midrule
    InternLM2-7B & 1 & 1600 tokens/s & 1$\times$ & 1.3M \\
    LLaMA3-70B & 4 & 720 tokens/s & 4.5$\times$ & 700K \\
    \bottomrule
    \end{tabular}%
    }
  \caption{Efficiency of using different LLM Writers for Input Generation. ``TP'' refers to the size of the tensor parallel for inference. ``TGS'' refers to the inference throughput (tokens/s) of each LLM instance. ``Costs'' refers to the relative compute costs of different LLM generators. Due to the large TP and low throughput, the large LMs can be more costly than the small LMs when inferencing on the same number of GPUs. In our data synthetic process, using LLaMA3-70B costs about 9$\times$ compute resources compared to small models like InternLM2-7B. Due to the high costs of LLaMA3-70B, we only sub-sample the raw data to run the data synthesis. The total costs are still 4.5$\times$ compared to InternLM2-7B.}
  \label{tab:generator-efficiency}%
\end{table}%

\begin{figure*}[htbp]
    \centering
    \includegraphics[width=0.8\textwidth]{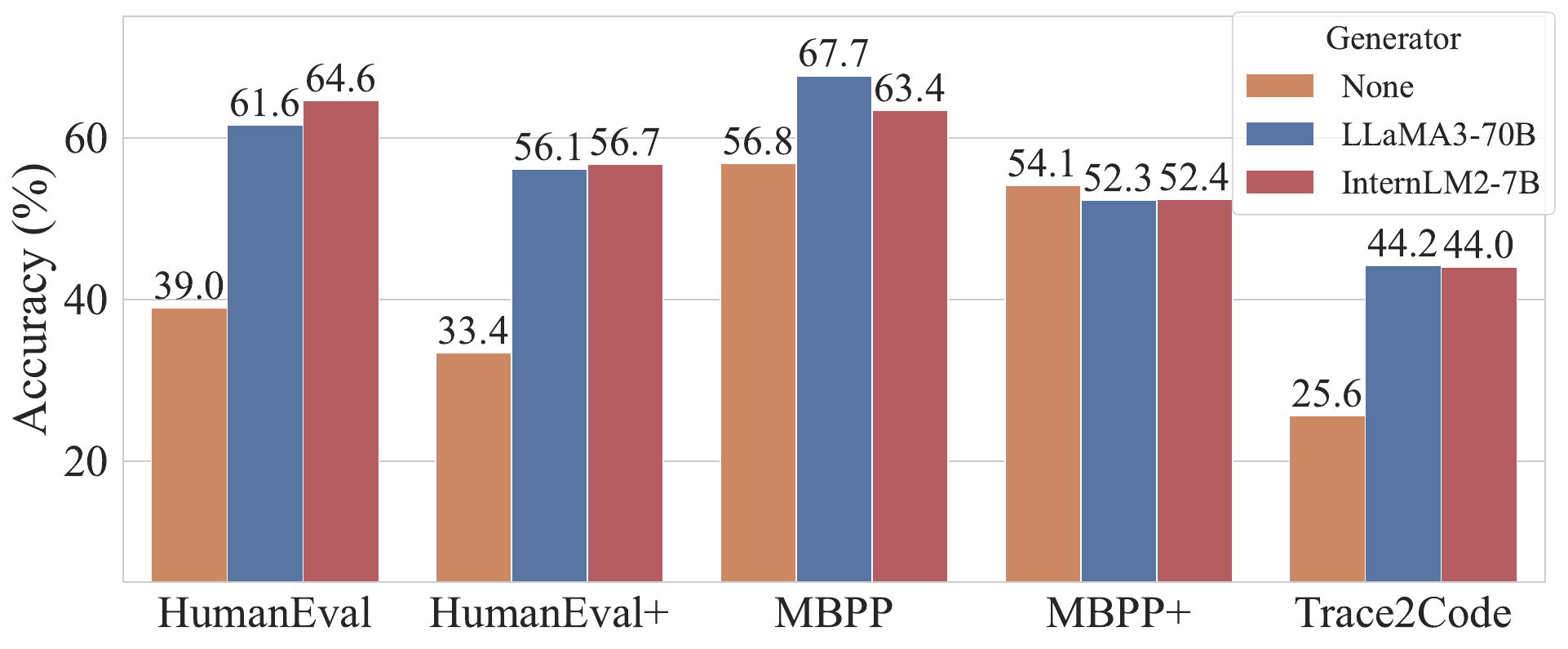}
    \caption{Downstream results when fine-tuning InternLM2-7B with synthetic data using different input example generators. Generator ``None'' refers to the baseline InternLM2-7B not trained on any \mytask~data. The computational overhead of using LLaMA3-70B is $4.5\times$ that of InternLM2-7B.}
    \label{fig:generator-ablation}
\end{figure*}
\begin{table*}[h!]
  \centering
    \begin{tabular}{l|cccc|c}
    \toprule
      & HumanEval & HumanEval+ & MBPP & MBPP+ & Case2Code \\
    \midrule
    InternLM2-1.8B & 32.3 & 29.9 & 43.6 & 24.3 & 27.8 \\
    InternLM2-7B & 64.6 & 56.7 & 63.4 & 52.4 & 42.2 \\
    InternLM2-20B & \textbf{73.1} & \textbf{65.2} & \textbf{77.4} & \textbf{55.4} & \textbf{46.0} \\
    \bottomrule
    \end{tabular}%
  \caption{Code results with different scales of models, after supervised fine-tuning on the instruction-following dataset mixed with \mytask~synthetic data.}
  \label{tab:model-scale-ablation}%
\end{table*}%

\paragraph{Importance of I/O pairs}
We extract code data from the case2code training data and remove other parts like prompts form the training set, \mytask~(code only). Then, we train Llama3-8B under the same setups of the mixed pre-training in Section~\ref{sec:mixed-training}, replacing the \mytask~(full) training set with the \mytask~(code only) set. Compared with mixed pre-trained models with \mytask~(code only) dataset, the \mytask~(full) trained models achieve much higher accuracy both on in-distribution \mytask~task and out-of-distribution tasks like HumanEval and HumanEval+, demonstrating the effectiveness of combining the I/O messages with the corresponding code and the incorporation of synthetic case2code samples.

\paragraph{Choice of Input Generator}
\label{sec:generator-ablation}
During the synthesis of \mytask~data, a critical step is prompting the LLM to write several input examples for each program. These inputs are then executed with the corresponding programs one by one to obtain the program outputs, thus we can utilize these important contexts to construct \mytask~training data. To explore whether the reasoning ability of the LLM writer affects the synthetic data quality, we replace the LLM generator from Interlm2-7B to LLaMA3-70B, and rerun the data synthesis pipeline to obtain a new version of \mytask~training data. Due to the high costs of LLaMA3-70B, we only generate half the size of our original synthetic data. Detailed generation costs are reported in Table~\ref{tab:generator-efficiency}. We train Interlm2-7B with this version of \mytask~dataset under the instruction-tuning setup to evaluate the data quality. As shown in Figure~\ref{fig:generator-ablation}, compared with the InternLM2-7B generator, large LMs like LLaMA3-70B can write high-quality input samples that help trained LLMs to achieve comparable code reasoning capability with fewer training data. It indicates that the input generation step can affect the overall synthetic data quality, suggesting data collectors choose a strong LLM to be the input writer if compute resources are sufficient. However, we note that LLaMA3-70B contains too many parameters that are $4.5\times$ more costly than InternLM2-7B. By generating inputs with InternLM2-7B, our \mytask~data synthesis framework maintains generation efficiency and data quality. It also demonstrates the possibility of self-improving for LLMs on their code reasoning capabilities.

\paragraph{Improvements on Different Model Scale}
We want to explore whether the \mytask~data synthesized using a small model can still improve a large model, and how the model scale affects the learning process. Therefore, we use \mytask~data generated with InternLM2-7B to train models in the InternLM2 series to investigate these questions. The training is taken under the setting of data mixing with SFT dataset~\cite{luo2023wizardcoder} and the results are shown in Table~\ref{tab:model-scale-ablation}. Our synthetic data consistently enhances the code reasoning performance of various sizes of LLMs, even though one of the student models is almost three times larger than the model used for data synthesis. These results demonstrate the possibilities of weak-to-strong supervision in code-related tasks at scale.

\section{Conclusion}
We first construct a new benchmark \mytask~to evaluate the inductive reasoning capability of LLMs in the code domain. Then, we propose a data synthetic framework to construct \mytask~training samples at scale. By just using small LLMs and a code interpreter, we can collect high-quality \mytask~data from pre-training code texts automatically and efficiently. By training on various LLMs in multiple settings, we demonstrate the \mytask~can improve not only the inductive reasoning ability of LLM but also the general coding capabilities. We believe synthetic \mytask~is a promising way to continue improving the LLMs when human-generated data is exhausted.

\section*{Limitations}

In this work, we study Case2Code, a synthetic task for learning inductive reasoning capabilities. Our work is still limited in several aspects:
\begin{itemize}
    \item Potential harmful programs: we gather and filter programs from the pre-training code corpus, which excludes code that may contain dangerous operations like system calls, file manipulation, and network traffic that require careful safety checks and vulnerability mitigation. In the future one can incorporate a safe and reliable execution environment that supports these operations for Case2Code synthesis. 
    \item Programming languages: we focus on synthesizing Case2Code data using Python programs, as it is a commonly used programming language and can be easily and reliably manipulated and executed. Future work can extend the data synthesis framework to more programming languages and applications.
    \item Long context: some inputs or outputs of the given programs can be extremely long, which can be challenging to fit into the context window of current LLMs. Future work can explore efficient methods of representing and learning long-context case-to-code induction.
    \item Data modality: we represent cases in our Case2Code data as texts for LLM training, however, real-world programs often interact with multi-modal inputs and outputs like audio, image, and video. How to effectively collect and learn multi-modal inductive reasoning remains a big challenge.
\end{itemize}


\section*{Acknowledgments}
This work was supported by the National Natural Science Foundation of China (No. 62236004).


\bibliography{custom}
\appendix

\section{Prompts Used in \mytask}
We demonstrate the prompts used during the \mytask~synthesis, training, and evaluation as follows:
\begin{itemize}
    \item The prompt template for evaluating zero-shot \mytask~performance of various LLMs is listed in Table~\ref{tab:appendix-zs-evaluation-prompt}.
    \item We show the prompt for using LLMs as input generators for synthesizing \mytask~data in Table~\ref{tab:appendix-generator-prompt}.
    \item We randomly sample some \mytask~ data to demonstrate in Table~\ref{tab:appendix-data-examples}.
\end{itemize}

\begin{table*}[htbp]
    \centering
    \small
    \begin{tabular}{l}
    \toprule
    \textbf{Prompt Template for Zero-shot \mytask~Evaluation.} \\
    \midrule
    \begin{minipage}{0.8\textwidth}
\begin{spverbatim}
{prompt}

Please write the correct names of arguments. As the function you implement will be called by: {func_name}(**input_dict). Keep the original type. No need to convert the output to string.
\end{spverbatim}
        \end{minipage}\\
    \bottomrule
    \end{tabular}
    \caption{Prompt template for zero-shot \mytask~evaluation. We inject \texttt{\{prompt\}} and \texttt{\{func\_name\}} for each test sample for evaluation.}
    \label{tab:appendix-zs-evaluation-prompt}
\end{table*}

\label{sec:appendix}
\begin{table*}[htbp]
    \centering
    \small
    \begin{tabular}{l}
    \toprule
    \textbf{Prompt for LLM Input Generator} \\
    \midrule
    \begin{minipage}{0.8\textwidth}
\begin{spverbatim}
Given the function, first analyze the types of the function arguments, then write 10 different example inputs for the function, each example should be a dict with function arguments' names and their values.
Output format:
```python
examples = [
    dict(argname=argvalue),
    ....
]
```

Function:
```python
def test_func(a: int, b: str) -> str:
    return str(a) + b
```
Examples:
```python
examples = [
    dict(a=1, b='a'),
    dict(a=2, b='b'),
    dict(a=3, b='c'),
    dict(a=4, b='d'),
    dict(a=5, b='e'),
    dict(a=6, b='f'),
    dict(a=7, b='g'),
    dict(a=8, b='h'),
    dict(a=9, b='i'),
    dict(a=10, b='j'),
]
```

Function:
```python
{code}
```
Examples:
\end{spverbatim}
        \end{minipage}\\
    \bottomrule
    \end{tabular}
    \caption{Prompt for LLM input generator, we replace \texttt{\{code\}} with programs collected in for \mytask.}
    \label{tab:appendix-generator-prompt}
\end{table*}

\begin{table*}[htbp]
    \centering
    \small
    \begin{tabular}{l}
    \toprule
    \textbf{\mytask~Examples} \\
    \midrule
    \begin{minipage}{0.8\textwidth}
\begin{spverbatim}
Input: dict(s="abcba", center=2), Output: "(5, 0, 4)"
Input: dict(s="abcdefg", center=3), Output: "(1, 3, 3)"
Input: dict(s="aba", center=1), Output: "(3, 0, 2)"
Input: dict(s="racecar", center=3), Output: "(7, 0, 6)"
Input: dict(s="madam", center=2), Output: "(5, 0, 4)"
Input: dict(s="abcabcabc", center=4), Output: "(1, 4, 4)"
Input: dict(s="xyzyx", center=2), Output: "(5, 0, 4)"
Input: dict(s="hello", center=2), Output: "(1, 2, 2)"
Input: dict(s="ab", center=0), Output: "(1, 0, 0)"
Input: dict(s="a", center=0), Output: "(1, 0, 0)"

Write a function that generates the output from the input.
Function: greatest_palindrome_size_odd
\end{spverbatim}
\end{minipage}\\ \\
\begin{minipage}{0.8\textwidth}\begin{spverbatim}
The function is:

```python
def greatest_palindrome_size_odd(s, center):
    right = center + 1
    left = center - 1
    size = 1
    optimum_right = optimum_left = center
    while left >= 0 and right < len(s):
        if s[left] == s[right]:
            size += 2
            optimum_left = left
            optimum_right = right
            right += 1
            left -= 1
        else:
            break
    return size, optimum_left, optimum_right

```
\end{spverbatim}\end{minipage}\\
\midrule
\begin{minipage}{0.8\textwidth}\begin{spverbatim}
Arguments and results:
Input: dict(seq="ATCG", complementarity={'A': 'U', 'T': 'A', 'C': 'G', 'G': 'C'}), Output: "'CGAU'"
Input: "ATCG", {'A': 'T', 'T': 'A', 'C': 'G', 'G': 'C'}, Output: "'CGAT'"
Input: seq:"ACGT", complementarity:{'A': 'U', 'T': 'A', 'C': 'G', 'G': 'C'}, Output: "'ACGU'"
Input: "ACGT", {'A': 'T', 'T': 'A', 'C': 'G', 'G': 'C'}, Output: "'ACGT'"

Please write a function to process the input arguments and produce the specified outputs.

Start with the function:
reverse_complement

The function is:

```python
def reverse_complement(seq, complementarity):
    bases = list(seq)
    bases = [complementarity[base] for base in bases]
    reversed_complement = ''.join(bases)
    return reversed_complement[::-1]

```
\end{spverbatim}\end{minipage}\\

    \bottomrule
    \end{tabular}
    \caption{\mytask~data examples.}
    \label{tab:appendix-data-examples}
\end{table*}

\end{document}